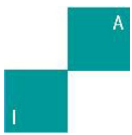

# INTELIGENCIA ARTIFICIAL

## Performance of Deep learning models with transfer learning for multiple-step-ahead forecasts in monthly time series

Martín Solís[1], Luis-Alexander Calvo-Valverde[2],

[1] Tecnológico de Costa Rica, Cartago, 159-7050, Costa Rica, marsolis@itcr.ac.cr
[2] Tecnológico de Costa Rica, Computer Science School, Cartago, 159-7050, Costa Rica, lcalvo@itcr.ac.cr

**Abstract** Deep learning and transfer learning models are being used to generate time series forecasts; however, there is scarce evidence about their performance prediction mainly for monthly time series. The purpose of this paper is to compare deep learning models with transfer learning and without transfer learning and other traditional methods used for monthly forecasts to answer three questions about the suitability of deep learning and transfer learning to generate predictions of time series. Time series of M4 and M3 competitions were used for the experiments. The results suggest that deep learning models based on tcn, lstm, and cnn with transfer learning tend to surpass the performance prediction of other traditional methods. On the other hand, tcn and lstm, trained directly on the target time series, got similar or better performance than traditional methods for some forecast horizons.

**Resumen** Los modelos de aprendizaje profundo y aprendizaje por transferencia se están utilizando para generar pronósticos de series temporales; sin embargo, existe escasa evidencia sobre su desempeño predictivo principalmente en series mensuales. El propósito de este artículo es comparar modelos de aprendizaje profundo con aprendizaje de transferencia y sin aprendizaje de transferencia y otros métodos tradicionales utilizados para pronósticos mensuales para responder tres preguntas sobre la idoneidad del aprendizaje por transferencia para generar predicciones de series temporales. Se utilizaron series temporales de las competiciones M4 y M3 para los experimentos. Los resultados sugieren que los modelos de aprendizaje profundo basados en tcn, lstm y cnn con aprendizaje por transferencia tienden a superar la predicción de rendimiento de otros métodos tradicionales. Por otro lado, tcn y lstm, entrenados directamente en la serie temporal objetivo, obtuvieron un rendimiento similar o mejor que los métodos tradicionales para algunos horizontes de pronóstico.

## 1 Introduction

An increasing number of deep learning models for time series forecasting have been published in conferences and journals in recent years [1]. These models have been successful for predictions in different areas [2] and, in some cases, yielded a superior predicted performance than traditional techniques [3]. For example, Zeroual et al [4] showed the promising potential of the deep learning model in forecasting COVID-19. Mariano-Hernández [5] found that deep learning models were the best to predict energy consumption in one of the buildings analyzed in their research. Hewage et al [6] showed that deep learning models could be better than the classic machine learning and statistical approaches to predict weather conditions.

Transfer learning is a technique used successfully in computer vision and NLP to extract the knowledge from a source domain and used in learning a model on a target domain [7]. Although transfer learning has not been widely used for deep learning models of time series [8], its application has emerged recently. For example, some authors have used transfer learning to deal with the scarcity of labeled time-series of data in classification cases [e.g 9,10]. Qi-Qiao et al [8] used transfer learning to make predictions with financial time series extracted from stock markets. Otovi et al [11] and Poghosyan et al [12] proved that a pre-trained model on a global time-series database could be used with transfer learning to get good predictions of other times series that even come from a different domain.

Due to the relevance that deep learning and transfer learning models have acquired for time series forecasting, there is a need to benchmark them against more traditional methods [2]. Performance comparisons have been made using statistical forecasting methods and classical machine learning methods [13, 14 15, 16]. However, comparisons and analyses about the predictive performance between those methods and deep learning models are scarce. This research aims to contribute to the aforementioned knowledge gap, answering the following questions.

1. Are the deep learning models for multiple-step-ahead forecasts in monthly time series more effective in terms of performance prediction than traditional methods?
2. How does the performance of the deep learning models change according to the forecast horizon compared to the traditional models?
3. Are there groups of time series where deep learning methods and the application of transfer learning are more effective?

We hope that this work provides valuable information about the suitability of deep learning models and the usage of pre-trained models to make predictions with transfer learning on new target time series. The main contributions of this paper are as follows:

1. We answered three questions about the effectiveness of deep learning models for monthly forecasts not responded yet.
2. Although there are benchmarking about the time series methods for forecasting, our work is the first, as far as we know, to compare and analyze the performance of the deep learning models with and without transfer learning with other more traditional machine learning algorithms and statistical methods.
3. The performance of deep learning models with transfer learning for monthly forecasts is analyzed throughout relevant characteristics of time series. Therefore, this work can be used as a reference point about the suitability of applying deep learning models and transfer learning for monthly forecast tasks.

## 2  Related Work

### 2.1  Deep learning with transfer learning in time series

Laptev, Yu, and Rajagopal [17] transfer time-series features across diverse domains and mentioned that there is no prior work related to the application of transfer learning for time-series. From 2018 some studies have emerged with the application of transfer learning to deep learning models to generate the forecast for a target time series. Those models have been applied to different tasks as time series classification tasks [e.g. 9, 10, 18], time series regression tasks [11, 12] and anomy detection in time series [19, 20].

In the case of regression tasks (the scope of this research), the transfer learning has been used in different areas. In finance, Qi-Qiao et al [8] evaluate the effectiveness of transfer learning for stock price prediction using a two-layer neuronal network and two-layer lstm. Xu y Meng [21] developed a novel hybrid transfer learning model for energy consumption forecasting based on time series decomposition. Le et al, [22] were concerned with the computational time to train several models to predict the energy consumption of the apartments at the same building; therefore, they developed a framework for multiple electric energy consumption forecasting in smart buildings based on transfer learning. Due to the problem of developing predictive models with limited data for energy load and power generation, [23] propose a transfer learning strategy using a convolutional neuronal network. Karb et al, [24] were also concerned about making predictions with limited time series, but in the domain of food sales of new products. They propose a network-based transfer learning approach for deep neural networks to create effective predictive models.

For crude oil price forecasting, Xiao et al, [25] generate a hybrid transfer learning-based analog complexing model (HTLM). Otovic et al [11] analyzed if knowledge transfer between related domains could be more beneficial than knowledge transfer between unrelated domains in classification and regression time-series tasks. They used datasets from diverse areas such as seismic datasets, acoustic signals, medical datasets, and stock-market prices. Xin and Peng [26] combines autoencoders, convolutional neural networks (AE-CNN), and transfer learning to capture the intrinsic certainty of chaotic time series. The authors of all previous studies demonstrate that the models were better than the baselines. In some cases, the comparison was with other deep learning, and in others, cases were statistical models as arima. The conclusions often highlight the promising of transfer learning for time series.

## 2.2    Performance Comparisons between forecast methods

Some comparisons have recently been made between the performance of various algorithms and methods for time series forecasting. One of the most cited is the work of [16]. Those authors compare the performance of seven statistical models and ten machine learning models using the M3 competition. Due to high computational time, they applied deep learning  models such as the lstm and rnn, but without transfer learning and based on very simple architectures. They conclude that traditional statistical methods are more accurate than machine learning ones. Papacharalampous et al [15] also extensively compare several stochastic and ML techniques to forecast hydrological processes. The machine learning models used were:

- ✓   Three simple neuronal networks (Single hidden layer, Multilayer, and Perceptron (MLP)).
- ✓   Three random forest.
- ✓   Three support vector machines.

Different to [16], they conclude that the stochastic and ML methods can share a quite similar performance when forecasting hydrological time series of small length, but in linear situations, the ML methods are more likely to be inferior, while in non-linear situations, the ML methods are more likely to outperform.

Parmezan et al [14] provide a comparison between popular statistical methods and machine learning models (support vector machines (svm), kNN-TSPI, Multilayer and Perceptron (mlp), and lstm), using synthetic and real times series of different domains. They conclude that SARIMA is the best for deterministic series, kNN-TSPI and sarima are the best for Stochastic, and svm was the more stable method for chaotic series. Catal et al [13] also compared machine learning models (linear regression, bayesian regression, neural network regression, decision forest regression, and boosted decision tree regression) and statistical models using the Walmart sales dataset. They found that boosted decision tree regression algorithm was the best predictor.

On the other hand, Lara-Benitez et al, [27] generate an experimental study comparing the performance of the most popular deep learning architectures for time series prediction using different datasets. In this study, they didn't compare with other machine learning algorithms or statistical methods nor analyzed the effect of transfer learning on the performance prediction. lstm, cnn, and tcn were between networks with the best results.

# 3    Material and methods

## 3.1    Datasets

The time series used comes from the M3 [39] and M4 competitions [40]. Two sets called A and B, of 1000 time series, were selected randomly from the Monthly M4 set, which contains 48 000 time series, and a third set named B_M3 were composed by all-time series of monthly M3 competition. Every time-series taken from the M4 competition includes the train and test subset. The testing subset of M4 competition for each time series in the A set was used to calculate the models' performance metrics. This subset is made up of 18 months.

As the research questions require the analysis of multiple-step-ahead forecasts (forecast horizon) performance with different steps, we pre-process the 18 months to make predictions of one step ahead, three, six, and twelve steps ahead. For example, for three steps ahead, the 18 test months were transformed into sixteen instances of prediction (table 1).

Table 1. Testing dataset structure for
output, using Three steps ahead

| instance | Month | | |
|----------|-------|-------|-------|
|          | Step1 | Step2 | Step3 |
| 1        | 1     | 2     | 3     |
| 2        | 2     | 3     | 4     |
| 3        | 3     | 4     | 5     |

………………………………………
………………………………………
………………………………………

......................................

| 15 | 15 | 16 | 17 |
| 16 | 16 | 17 | 18 |

## 3.2    Experimental procedure

The machine learning and deep learning   models were trained using different input sizes for each forecast horizon (time steps ahead). For 3, 6, and 12 steps ahead, we applied a window input size of:

✓ the same size as the forecast horizon. This input size was chosen because Shynkevich et al [28] found that the highest prediction performance is achieved when the input window length is approximately equal to the forecast forecast.

✓ 1.25 times of the forecast horizon size. This input size was chosen because Lara-Benitez et al [27] found better performance with 1.25 times the forecast horizon size than larger output horizons.

✓ 12 months. This input size was chosen to incorporate stational information into the models.

The procedure for training and testing each of the methods used is explained below:

### 3.2.1    Deep learning models with transfer learning

Figure 1 describes the process. Source set was used to develop the models[1] and set A to make the transfer learning and get the predictions of the test sample[2]. There were two kinds of source datasets: B set (1000 time series selected randomly from M4 competition) and B_M3 set (monthly time series of M3 competition). The process executed was the next. First, the input window size and output horizon size were selected for the pre-processing of each time series in every set. Then the data is divided into training, validation, and testing samples. The train and validation samples of the all-time series in the source set were concatenated to train the models and select the hyper-parameters and architectures using Bayesian optimization. The grid of search for the Bayesian optimization and the most frequent solution (between all models generated for each combination of input-output) is in table 2.

Each model has been trained using the Adam optimizer and a stop criterion which consists of stopping after two epochs without improvement in the validation sample's loss function. The loss function was the mean absolute percentage error, and the batch size was equal to one. After the model has been generated, transfer learning is applied for every time series of set A (which is the target dataset).  For the transfer learning we executed the next steps with each time series in dataset A:

1. The training sample was used to update the weights of the last layer in the trained models with a learning rate of 0.000005. This learning rate is smaller than the one used to build the models. The rest of the weights were frozen.

2. After two epochs without improvement in the validation sample's loss function, the updating process finishes.

3. Finally, the predictions were generated with the model updated and performance metrics were computed using the test sample.

We generated two versions of the deep learning models that were applied with transfer learning. In the first version, we trained the cnn, lstm, and tcn using the B set, and in the second version, we trained these networks using the B_M3 set. The purpose of the two versions was to assess how transfer learning worked when the models had been trained from different datasets.

---

[1] Due to the Bayesian optimization process, the network architecture changes between the models generated for the different input-output combinations. In the next repository (https://github.com/martin12cr/DatasetsDeepTransfer.) is the code of the base architectures from which the final models could change.

[2] The datasets are available in the next repository  (https://github.com/martin12cr/DatasetsDeepTransfer). The test set for each time series in dataset A is composed by the last 18 points.

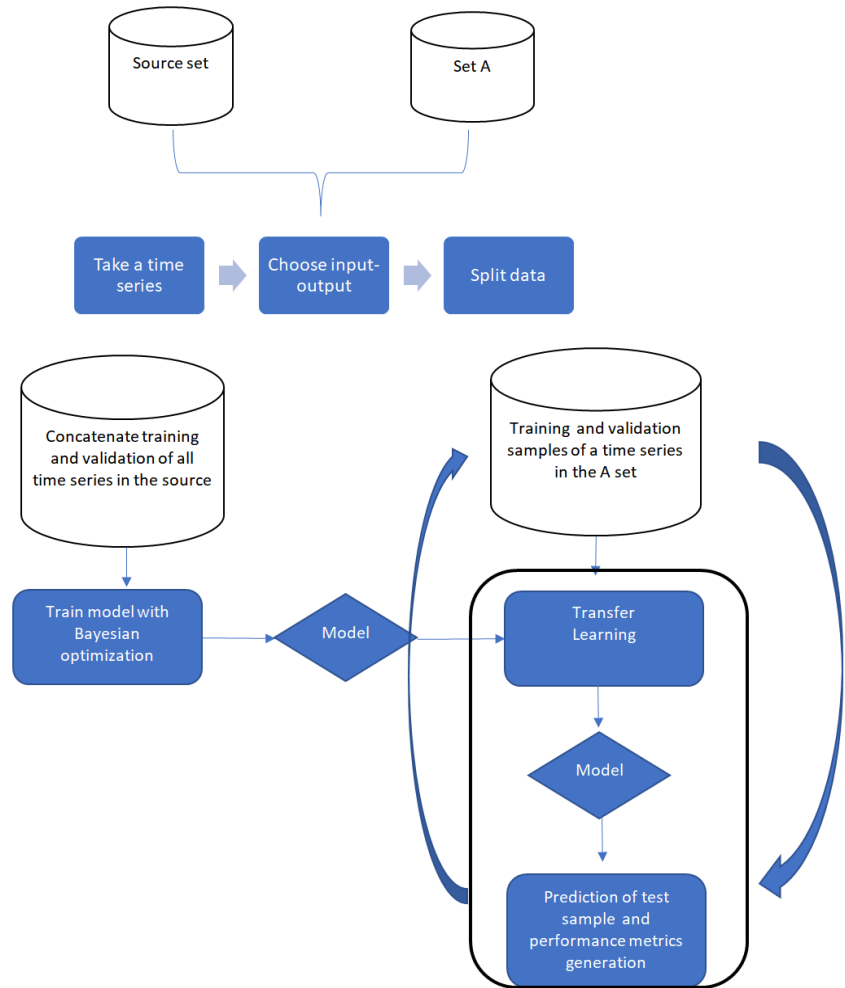

Figure 1. Procedure to train and test deep learning models with transfer learning

Table 2. Grid of search for the Bayesian optimization of deep learning models and most frequent solution between the models

| cnn | | tcn | | lstm | |
|---|---|---|---|---|---|
| search grid | Most frequent solution | search grid | Most frequent solution | search grid | Most frequent solution |
| ✓ Number of hidden layers between 1 and 2 (Batch normalization is applied after the first layer) | 2 | ✓ Filters between 12 and 132 with a step of 24 | 12 | ✓ Recurrent units between 12 and 132 with a step of 24 | 84 |
| ✓ Filters between 12 and 132 with a step of 24 | 12, both layers) | ✓ Kernel size between 2 and 12 with a step of 2 | 12 | ✓ Activation function among linear, relu, and tanh | Relu |
| ✓ Kernel size between 2 and 12 with a step of 2 | 12 (first) 2 and 12 (second) | ✓ Activation function among linear, relu, and tanh | Tanh | ✓ Return sequences True or False | True |
| | | ✓ Return sequences True or False | False | ✓ Learning rate among 0.001, 0.0001, 0.00001 | 0.001 |
| ✓ Max pooling with a size of 2 or without max pooling | Whithout | ✓ Learning rate among 0.001, 0.0001,0.00001 | 0.001 | | |
| ✓ Activation function among linear, relu, and tanh | Tanh | ✓ Dilations between [1,2,4,8] or [1,2,4,8,16] | [1,2,4,8,16] | | |
| ✓ Learning rate among 0.001, 0.0001,0.00001 | 0.0001 | | | | |

### 3.2.2    Machine learning and deep learning without transfer learning

Figure 2 describes the process. These models were developed with the training and validation samples of set A and evaluated with the test sample. The Bayesian optimization was used to select the hyperparameters of the models and the architectures of the deep learning    models. The grid of search is in table 2 for deep learning  models, and in table 3 for machine Learning models. When the model has been trained for each time series, the predictions and performance metrics were generated using the test dataset.

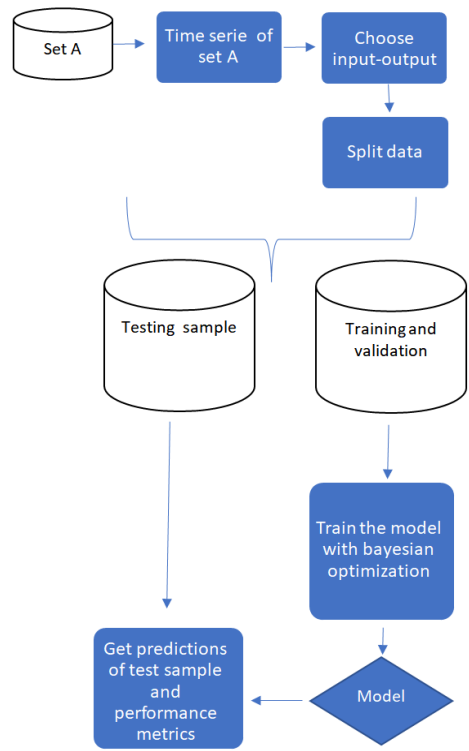

Figure 2. Procedure to train and test Machine Learning models

Table 3. Grid of search for the Bayesian optimization of machine learning models and most frequent solution between the models

| xgboost | | support vector machines | | random forest | |
|---|---|---|---|---|---|
| search grid | Most frequent solution | search grid | Most frequent solution | search grid | Most frequent solution |
| ✓ Max depth between 2 and 12 | 4<br><br>0.4-0.5 | ✓ C between 0.01 and 10 | 9.97-10 | ✓ estimators between 10 and 250 | 110-135 |
| ✓ Learning rate between 0.01 and 1 | | ✓ Gamma between 0.001 and 0.33 | 0.0001 - 0.05 | ✓ max_features between 1 and 15] | 2 |
| ✓ estimators between 10 and 150 | 70-79 | ✓ Kernel=rvf | | ✓ min sample leaf between 1 and 8 | 1 |
| | | | | ✓ mas samples between 0.70 and 0.99 | 0.70-0.75 |

### 3.2.3    Statistical models

The methods used were auto arima, ets, theta. The training and testing process is similar to figure 2 but without Bayesian Optimization component, because these statistical methods use other techniques to define the model function.

## 3.3    Performance metrics

In order to evaluate the performance of the models, two metrics were estimated using the test sample, mape and sMape. We chose these metrics because they allow the performance comparison between time series independently of the scale. The time series were normalized only for the deep learning    models, but the metrics were estimated on the original scale after converting the predictions to the original scale.

$$\text{Mape} = \frac{1}{n}\sum_{t=1}^{n}\left|\frac{y_t - \hat{y}_t}{y_t}\right| \qquad\qquad \text{sMape} = \frac{1}{n}\sum_{t=1}^{n}\left|\frac{y_t - \hat{y}_t}{|y_t| + |\hat{y}_t|}\right|$$

## 3.4    Software

The models and the experimental procedure were developed in python 3.6. The deep learning models were developed and trained with keras, the machine learning models were trained with scikit-learn and the statistical models were generated with sktime. The bayes_opt library was used for the optimization process of the machine learning models and keras library for the optimization process of the deep learning models.

# 4    Results

## 4.1    4.1. Best input size

Table 4 shows the relative distribution of the best input size for the all-time series in the target dataset according to the type of model and output horizon. The deep learning models with transfer learning tend to get the best results with an input of 12 than deep learning models without transfer learning and machine learning models, for outputs horizons of 1,3 and 6. For an output of 12, the best option in most time series was an input of 1.25 times the size of the output horizon, in deep learning and machine learning models. These results suggest that the best input depends on the time series and that there is a tendency for some types of models and output horizons to get the best results with higher inputs.

Table 4. Relative distribution of the best input size according to output horizon and type of model.

| Output | Input | deep learning_M4 | deep learning_M3 | deep learning sin transfer | machine learning |
|--------|-------|------------------|------------------|----------------------------|------------------|
|        | 3     | 41.9%            | 28.8%            | 59.3%                      | 55.1%            |
|        | 12    | 58.1%            | 71.2%            | 40.7%                      | 44.9%            |
| 1      | Total | 100.0%           | 100.0%           | 100.0%                     | 100.0%           |
|        | 3     | 32.0%            | 19.1%            | 35.3%                      | 31.0%            |
|        | 4     | 24.2%            | 23.2%            | 31.8%                      | 32.9%            |
|        | 12    | 43.7%            | 57.8%            | 32.9%                      | 36.1%            |
| 3      | Total | 100.0%           | 100.0%           | 100.0%                     | 100.0%           |
|        | 6     | 25.2%            | 25.0%            | 34.1%                      | 42.8%            |
|        | 8     | 22.9%            | 19.6%            | 33.7%                      | 23.1%            |
|        | 12    | 51.9%            | 55.3%            | 32.2%                      | 34.2%            |
| 6      | Total | 100.0%           | 100.0%           | 100.0%                     | 100.0%           |
|        | 12    | 45.1%            | 38.0%            | 52.7%                      | 44.5%            |
|        | 15    | 54.9%            | 62.0%            | 47.3%                      | 55.5%            |
| 12     | Total | 100.0%           | 100.0%           | 100.0%                     | 100.0%           |

## 4.2    Performance by forecast horizon

Table 5 shows the performance for each model and forecast horizon, based on mape and sMape. The machine learning models and deep learning   models were trained with different input window sizes for each output horizon; however, we selected the models with the input size that let to obtain the best performance.

According to the results, the statistical models have more stable performance through the forecast horizons; meanwhile, the machine learning models show an increment as the forecast horizon increases. The deep learning models with transfer learning tend to show the better performance at the mape and sMape in each forecast horizon, but to determine if there are significant statistical differences between models, the CD Diagrams were generated, following the procedure of the package autorank from python [29].

In the CD diagram, the vertical lines show the average rank of each model compared to the others. This average is calculated based on the performance metric used for the test sample. The horizontal line represents the critical difference for the comparison between models. When the average distance between models is greater than the critical distance, there is a statistically significant difference in the performance using a p-value =0.05. Therefore, when the horizontal line intersects the vertical lines of the average rankings, there is no difference between models.

Table 5. Mape and sMape on test by type of model and forecast horizon

Mape                                                                    sMape

| Model | Forecast horizon | | | |
|---|---|---|---|---|
| | 1 | 3 | 6 | 12 |
| arima | 0.199 | 0.192 | 0.195 | 0.206 |
| ets | 0.167 | 0.17 | 0.169 | 0.169 |
| theta | 0.159 | 0.16 | 0.159 | 0.156 |
| rf | 0.136 | 0.13 | 0.154 | 0.238 |
| svm | 0.136 | 0.129 | 0.148 | 0.241 |
| xgb | 0.136 | 0.129 | 0.148 | 0.254 |
| cnn_wot | 0.138 | 0.155 | 0.179 | 0.219 |
| lstm_wot | 0.107 | 0.115 | 0.136 | 0.167 |
| tcn_wot | 0.106 | 0.123 | 0.155 | 0.2 |
| cnn | 0.093 | 0.112 | 0.123 | 0.14 |
| lstm | 0.091 | 0.097 | 0.111 | 0.131 |
| tcn | 0.087 | 0.098 | 0.111 | 0.131 |
| cnn2 | 0.094 | 0.1 | 0.115 | 0.134 |
| lstm2 | 0.09 | 0.101 | 0.114 | 0.133 |
| tcn2 | 0.088 | 0.101 | 0.116 | 0.139 |

| Model | Forecast horizon | | | |
|---|---|---|---|---|
| | 1 | 3 | 6 | 12 |
| arima | 0.073 | 0.073 | 0.074 | 0.077 |
| ets | 0.064 | 0.066 | 0.066 | 0.065 |
| theta | 0.062 | 0.063 | 0.063 | 0.062 |
| rf | 0.057 | 0.054 | 0.064 | 0.107 |
| svm | 0.057 | 0.054 | 0.062 | 0.107 |
| xgb | 0.056 | 0.054 | 0.062 | 0.108 |
| cnn_wot | 0.053 | 0.07 | 0.08 | 0.093 |
| lstm_wot | 0.045 | 0.05 | 0.057 | 0.069 |
| tcn_wot | 0.044 | 0.053 | 0.065 | 0.086 |
| cnn | 0.038 | 0.048 | 0.051 | 0.056 |
| lstm | 0.037 | 0.04 | 0.045 | 0.052 |
| tcn | 0.036 | 0.041 | 0.045 | 0.052 |
| cnn2 | 0.039 | 0.042 | 0.047 | 0.053 |
| lstm2 | 0.037 | 0.042 | 0.046 | 0.053 |
| tcn2 | 0.036 | 0.042 | 0.047 | 0.055 |

rf: Random forest, svm: Support vector machines, xgb: Xgboost, cnn_wot=Convolutional neuronal network without transfer learning, lstm_wot: Long short term memory without transfer learning, tcn_wot: Temporal convolutional network without transfer learning, cnn: Cnn with transfer learning, lstm: Lstm with transfer learning, tcn: Tcn with transfer learning: Cnn with transfer learning trained with M4 dataset, lstm: Lstm with transfer learning trained with M4 dataset, tcn: Tcn with transfer learning trained with M4 dataset, tcn2: Tcn with transfer learning trained with M3 dataset, lstm2: Lstm with transfer learning trained with M3 dataset, cnn2: Cnn with transfer learning trained with M3 dataset.

Figure 4 shows the sMape CD diagrams for each forecast horizon, and figure 5 the mape CD diagrams. The results are similar in both graphs. Based on the hypothesis test, the best models when the forecast horizons are 1 and 3 months were the tcn with transfer learning and lstm with transfer learning. The first and second place depends on whether we look at the mape o sMape figure. According to the statistical test, the best model for a forecast horizon of one and three is tcn and the best model for a forecast horizon of six and twelve is lstm. The second position tends to belong to lstm1 and lstm2 or tcn1 and tcn2. According to the statistical test, the second position is shared by two or more methods in two graphs. For a forecast horizon of six in figure 4, there is no statistical difference between tcn and lstm2 and for a forecast horizon of one in figure 4, there is no statistical difference between lstm, tcn2, lstm2, cnn, cnn2, and tcn_wot. Another finding from the statistical test is that the last position tends to belong to cnn_wot or arima.

When the forecast horizon is larger, the lstm with transfer learning was first, and tcn or lstm trained with M3 dataset was second. The CNN and tcn without transfer learning tend to be the worst option for a larger horizon. Interestingly, the deep learning models with transfer learning but trained with a different dataset show good results because they were located in the second subset of the best models in each forecast horizon. The M3 monthly dataset used to train these models is not very different from the monthly M4 dataset however shows some differences; for example, the M3 time series are less forecastable, trended, and linear [30]. This finding suggests that it is possible to get good results when the training dataset has differences from the test dataset while these differences are not large. Some deep learning models without transfer learning were in the third group of best performance when the forecast horizon is less than 12, despite the exposition to overfitting, because of monthly time series with few data points and models with more parameters. Finally, the machine learning models (not deep learning models) were neither the third-best group nor the worst group.

**F.horizon= 1**

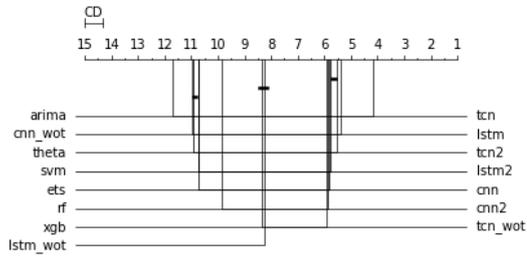

**F.horizon= 3**

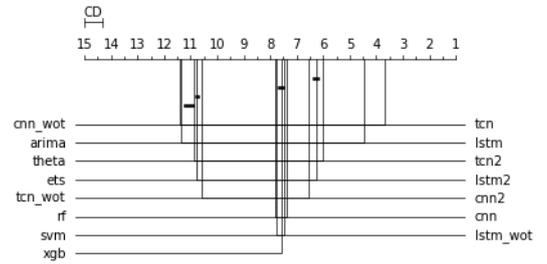

**F.horizon= 6**

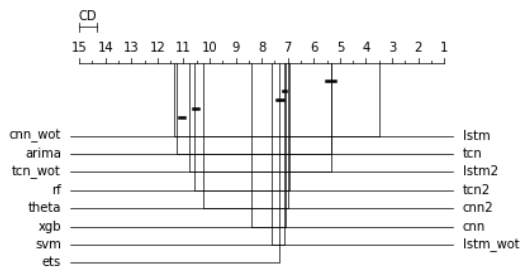

**F.horizon= 12**

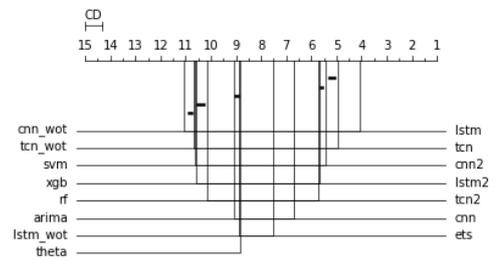

Figure 4. sMape CD diagrams for each forecast horizon.

**F.horizon= 1**

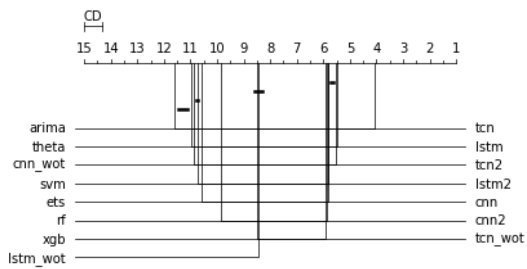

**F.horizon= 3**

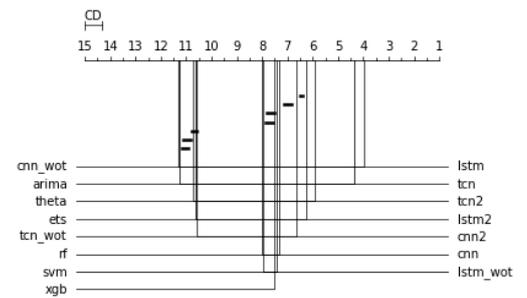

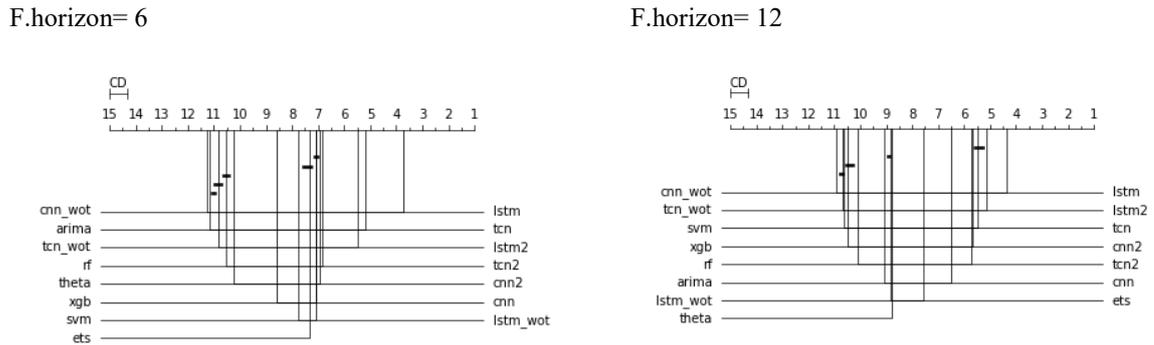

Figure 5. Mape CD diagrams for each forecast horizon.

## 4.3    Performance by type of time series

In order to analyze the performance of the models by the behavior of the time series, we estimated eight features for each time series, and later we applied the PAM cluster to classify the time series according to their behavior. The features calculated were:

1.   Forecastability measure with entropy [31]: Larger values occur when a time series is difficult to forecast.
2.   Seasonal [31]: Larger values means more seasonal strength
3.   Linearity/nonlinearity [31]: Takes large values when the series is non-linear and around 0 when the time series is linear.
4.   Skewness [31]: negative skew commonly indicates negative asymmetry distribution, positive skew indicates positive asymmetry distribution, and values close to zero indicates symmetry distribution
5.   Kurtosis [31]: Lager values means more concentration around the mean
6.   White noise measure with box test [32]. Higher values of chi-square test mean not white noise
7.   Outliers: Proportion of outliers computed using [33] approach
8.   Stationarity was measured with the adf test of stationarity [34]. Higher p values indicate that the series is not stationary

Before the PAM execution, the features were standardized. The silhouette score and calinski harabasz score were higher with two cluster; however, we decided to generate four to have more cluster diversity. Based on the means features of each cluster, we determine that the first and second clusters are composed for time series more unpredictable (see the entropy and white noise metrics), but in the second group, the time series tend to be more non-linear, asymmetric and with more concentration of data point around the mean. Clusters 3 and 4 are more predictable, but in 3, the time series tend to have more seasonal strength, while in group 4 tend to be more stationary. Another difference is that in 4, the time series tend to be more non-linear, with more outliers. At the bottom of table 2, we include brief names for each cluster based on the previous analysis.

Table 6. Features means by cluster

| Features | cluster1 | cluster2 | cluster3 | cluster4 |
|----------|----------|----------|----------|----------|
| entropy | 0.614 | 0.598 | 0.227 | 0.231 |
| season | 0.634 | 0.214 | 0.653 | 0.252 |
| skewness | 0.059 | 0.670 | 0.058 | 0.389 |
| kurtosis | 0.159 | 1.704 | -0.761 | -0.598 |
| non_linear | 0.338 | 0.666 | 0.235 | 0.676 |
| white_noise | 732.5 | 793.1 | 3458.0 | 2008.8 |
| outliers | 0.009 | 0.015 | 0.007 | 0.034 |
| stacionarity | 0.519 | 0.401 | 0.382 | 0.646 |

Cluster 1: unpredictable -symmetric, Cluster 2: unpredictable-nonlinear
with concentration around mean, Cluster 3: predictable-seasonable
Cluster 4: predictable - with stationary and nonlinear tendency

Figure 6 and Figure 7 presents the CD diagrams by series cluster, using sMape and mape, respectively.   The main finding in both figures is that the tcn and lstm with transfer learning were between the two best models.

According to the statistical differences, the best model for the cluster unpredictable -symmetric is tcn using the sMape and lstm2 using the mape. For unpredictable-nonlinear the best model is lstm, and in the case of predictable-seasonable and predictable - with stationary and nonlinear tendency the best models with both metrics are lstm and tcn, respectively. The statistical test algo suggest that the worst method is cnn_wot in five graphs and arima in three graphs. There are several methods that don't show statistical difference in positions three, four, five and six.

Also, the ranking of these figures shows that the deep learning models trained with a different dataset were frequently between the third and five positions. These results confirm that the performance of the deep learning models with transfer learning was between the best in different groups of series, that goes from the more predictable to the more unpredictable and noisy time series. The CNN without transfer learning was in the last position, and the tcn without transfer learning was not between the best seven models independently of the cluster. The lstm without transfer learning got the best results, but in some clusters like the unpredictable-nonlinear with a concentration around mean and predictable - with stationary and nonlinear tendency were surpassed by theta or ets.

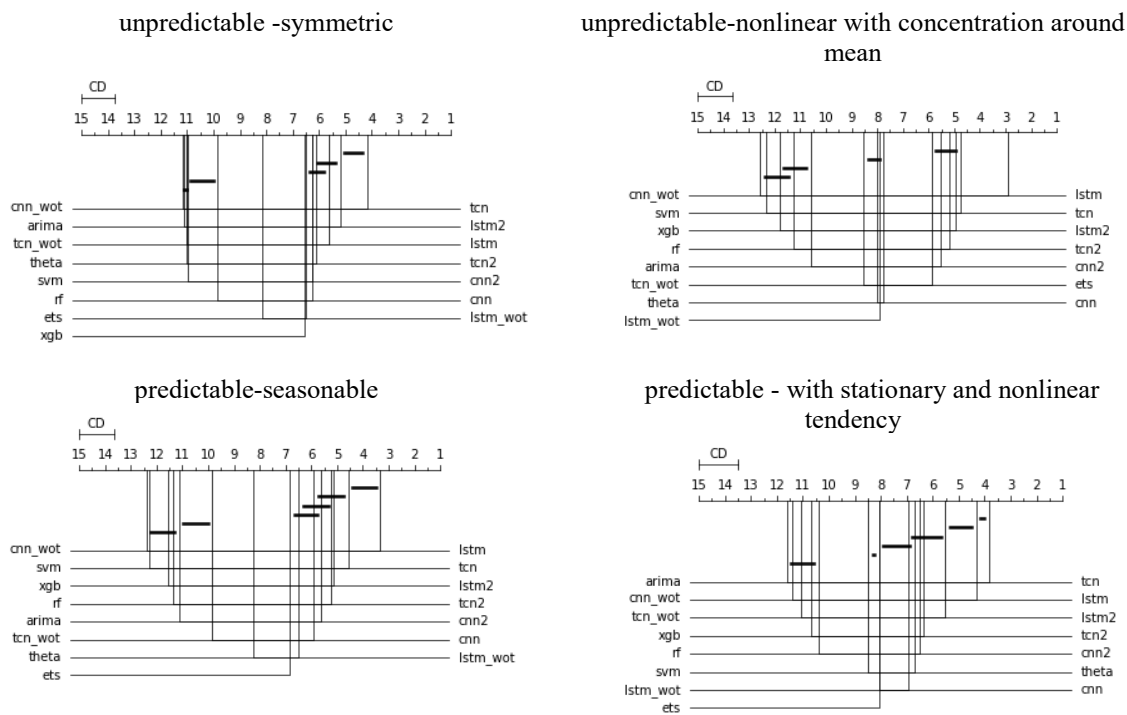

Figure 6. sMape CD diagrams for each cluster of series

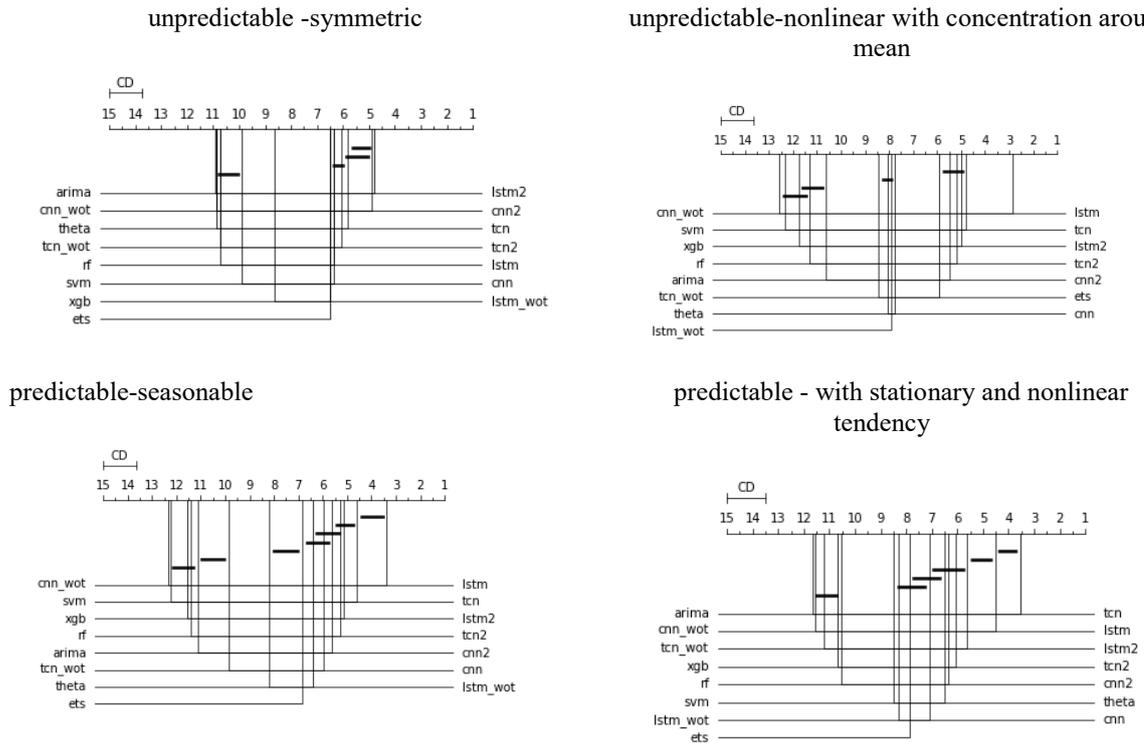

Figure 7. Mape CD diagrams for each cluster of series

## 5 Conclusions

Three research questions were posited in the introduction; they will be answered in this section. The first question was: Are the deep learning models for multiple-step-ahead forecasts in monthly time series more effective in terms of performance prediction than traditional models?

Based on the mape and sMape metrics, the deep learning models tcn, lstm, and cnn with transfer learning tend to be more effective than some more traditional models such as theta, ets, arima, random forest xgboost, and svm. Those deep learning models were trained with a concatenated dataset of time series, and later, the transfer learning was applied to update the weights of the last layer for new time series. Also, the models trained with the M3 dataset show the best performance than traditional methods. The monthly M3 dataset has some similarities with the monthly M4 dataset but is not entirely identical because the time series tends to be less forecastable, trended, and linear [30]. This finding suggests that the deep learning algorithms are a good option or maybe the best option to build forecast models if we make a dataset with time series that comes from a population that is the same or at least a little different from the population where our target time series comes from.

On the other hand, if we train the tcn, lstm, and cnn directly on the target time series, the results suggest that the lstm and tcn could get similar or better performance than traditional methods, depending on the forecast horizon, while the cnn is not a good option. It is reasonable that the deep learning models have lower performance being trained directly on the monthly target time series because this kind of series didn't have a lot of data points, and deep learning models require to estimate several weights, which could cause overfitting.

The second question was: How does the performance of the deep learning models change according to the forecast horizon compared to the traditional models?

The answer is that the statistical models are more stable in the performance if the forecast horizon increases, however the deep learning models with transfer learning show a best average ranking position and less mape and sMape independently if the forecast horizon is 1, 3, 6, or 12. According to the forecast horizon, the best option is lstm or tcn; for example, the results suggest that if the goal is to predict 1 or 3 months, the best option is tcn, but for larger horizons, the best is lstm. The cnn with transfer learning tends to have lower performance on any horizon.

The third question was: Are there groups of time series where deep learning methods and the application of transfer learning are more effective?

We classify the time series in four groups: unpredictable -symmetric, unpredictable-nonlinear with a concentration around mean, predictable-seasonable and predictable - with stationary and nonlinear tendency. Our results suggest that the deep learning models with transfer learning tend to be the best independently of the group; however, it doesn't mean that they are always the best option. The best deep learning model using transfer learning belongs to the tcn or lstm model in the four groups.

In relation with the deep learning models trained directly on the target time series, the result suggests that the lstm tends to have better performance than other traditional methods when the time series is unpredictable - symmetric and predictable-seasonable. In other groups, the ets or theta got better performance.

*Future research lines*

In this study, the performance of deep learning models with transfer learning was evaluated on time series that belong to a similar or slightly different population of the time series used to train the models. However, it is relevant to know how the transfer learning models would work as the distance between the features of the training and testing time series are further apart.

Clusters of time series have been generated to compare the performance of the algorithms by groups; however, the final goal should be to develop a meta-learning model that determines which algorithm would be the best according to the features of each time series. Some efforts have been made on this research line [e.g, 35, 36, 37, 38] without incorporating the deep learning models.

Last, our study didn't include the performance comparison of the deep learning models with hybrid models that combine different algorithms. Several studies have shown that this kind of models perform well in various topics.

In the field of Natural Language Process, there are pre-trained transformer models that have been used successfully to fine-tune a model on different tasks. For example, a pre-trained mask language model can be used to generate a model for text classification or text summarization. According to [41] "in time series there are limited works on pre-trained transformers and existing studies mainly focus on time series classification". Therefore, new studies should generate and evaluate the performance of pre-trained transformers for time series forecasting to determine in what conditions the transformers are more effective than other kinds of models. Additionally, it is relevant to explore different architectures of transformers not only the vanilla transformer. In [41] the authors found that different alterations to the vanilla transformer could bring good performance in time series forecasting.

## Acknowledgements

This work was supported by Tecnológico de Costa Rica.

## References

[1] Sezer, O. B., Gudelek, M. U., & Ozbayoglu, A. M. (2020). Financial time series forecasting with Deep learning : A systematic literature review: 2005–2019. Applied Soft Computing, 90, 106181. doi:10.1016/j.asoc.2020.106181

[2] Sharma, A., & Jain, S. K. (2021). Deep learning Approaches to Time Series Forecasting. R cent Advances i n Time Series Forecasting, 91–97. doi:10.1201/9781003102281-6

[3] Gamboa, J. C. B. (2017). Deep learning for time-series analysis. arXiv preprint arXiv:1701.01887.

[4] Zeroual, A., Harrou, F., Dairi, A., & Sun, Y. (2020). Deep learning methods for forecasting COVID-19 time-Series data: A Comparative study. Chaos, Solitons & Fractals, 140, 110121. doi:10.1016/j.chaos.2020.110121

[5] Mariano-Hernández, D., Hernández-Callejo, L., Solís, M., Zorita-Lamadrid, A., Duque-Perez, O., Gonzalez-Morales, L., & Santos-García, F. (2021). A Data-Driven Forecasting Strategy to Predict Continuous Hourly Energy Demand in Smart Buildings. Applied Sciences, 11(17), 7886. doi:10.3390/app11177886

[6] Hewage, P., Behera, A., Trovati, M., Pereira, E., Ghahremani, M., Palmieri, F., & Liu, Y. (2020). Temporal convolutional neural (TCN) network for an effective weather forecasting using time-series data from the local weather station. Soft Computing, 24(21), 16453–16482. doi:10.1007/s00500-020-04954-0

[7] Pan, S. J., & Yang, Q. (2009). A survey on transfer learning. *IEEE Transactions on knowledge and data engineering, 22*(10), 1345-1359.

[8] Qi-Qiao, H., Cheong-Iao, P., Yain-Whar, S. (2019, August). Transfer learning for financial time series forecasting. In *Pacific Rim International Conference on Artificial Intelligence* (pp. 24-36). Springer, Cham.

[9] Li, F., Shirahama, K., Nisar, M. A., Huang, X., & Grzegorzek, M. (2020). Deep Transfer learning for Time Series Data Based on Sensor Modality Classification. Sensors, 20(15), 4271. doi:10.3390/s20154271

[10] Gupta, P., Malhotra, P., Narwariya, J., Vig, L., & Shroff, G. (2019). Transfer learning for Clinical Time Series Analysis Using Deep Neural Networks. Journal of Healthcare Informatics Research, 4(2), 112–137. doi:10.1007/s41666-019-00062-3

[11] Otović, E., Njirjak, M., Jozinović, D., Mauša, G., Michelini, A., & Štajduhar, I. (2022). Intra- domain and cross-domain transfer learning for time series data—How transferable are the features? Knowledge-Based Systems, 239, 107976. doi:10.1016/j.knosys.2021.107976

[12] Poghosyan, A., Harutyunyan, A., Grigoryan, N., Pang, C., Oganesyan, G., Ghazaryan, S., & Hovhannisyan, N. (2021). An Enterprise Time Series Forecasting System for Cloud Applications Using Transfer learning. doi:10.20944/preprints202101.0326.v1

[13] Catal, C., Ece, K., Arslan, B., & Akbulut, A. (2019). Benchmarking of Regression Algorithms and Time Series Analysis Techniques for Sales Forecasting. Balkan Journal of Electrical and Computer Engineering, 20–26. doi:10.17694/bajece.494920

[14] Parmezan, A. R. S., Souza, V. M. A., & Batista, G. E. A. P. A. (2019). Evaluation of statistical  and machine learning models for time series prediction: Identifying the state-of-the-art and the best conditions for the use of each model. Information Sciences, 484, 302–337. doi:10.1016/j.ins.2019.01.076

[15] Papacharalampous, G., Tyralis, H., & Koutsoyiannis, D. (2019). Comparison of stochastic and machine learning methods for multi-step ahead forecasting of hydrological processes. Stochastic Environmental Research and Risk Assessment, 33(2), 481–514. doi:10.1007/s00477-018-1638-6

[16] Makridakis, S., Spiliotis, E., & Assimakopoulos, V. (2018). Statistical and Machine Learning forecasting methods: Concerns and ways forward. PLOS ONE, 13(3), e0194889. doi:10.1371/journal.pone.0194889

[17] Laptev, N., Yu, J., & Rajagopal, R. (2018, August). Reconstruction and regression loss for time-series transfer learning. In *Proceedings of the Special Interest Group on Knowledge Discovery and Data Mining (SIGKDD) and the 4th Workshop on the Mining and LEarning from Time Series (MiLeTS), London, UK* (Vol. 20).

[18] Nguyen, T.-T., & Yoon, S. (2019). A Novel Approach to Short-Term Stock Price Movement Prediction using Transfer learning. Applied Sciences, 9(22), 4745. doi:10.3390/app9224745

[19] Wen, T., & Keyes, R. (2019). Time series anomaly detection using convolutional neural networks and transfer learning. *arXiv preprint arXiv:1905.13628*.

[20] Xiong, P., Zhu, Y., Sun, Z., Cao, Z., Wang, M., Zheng, Y., … Que, Z. (2018). Application of Transfer learning in Continuous Time Series for Anomaly Detection in Commercial Aircraft Flight Data. 2018 IEEE International Conference on Smart Cloud (SmartCloud). doi:10.1109/smartcloud.2018.00011

[21] Xu, X., & Meng, Z. (2020). A hybrid transfer learning model for short-term electric load forecasting. Electrical Engineering, 102(3), 1371–1381. doi:10.1007/s00202-020-00930-x

[22] Le, T., Vo, M. T., Kieu, T., Hwang, E., Rho, S., & Baik, S. W. (2020). Multiple Electric Energy Consumption Forecasting Using a Cluster-Based Strategy for Transfer learning in Smart Building. Sensors, 20(9), 2668. doi:10.3390/s20092668

[23] Hooshmand, A., & Sharma, R. (2019). Energy Predictive Models with Limited Data using Transfer learning. Proceedings of the Tenth ACM International Conference on Future Energy Systems. doi:10.1145/3307772.3328284

[24] Karb, T., Kühl, N., Hirt, R., & Glivici-Cotruta, V. (2020). A network-based transfer learning approach to improve sales forecasting of new products. *arXiv preprint arXiv:2005.06978*.

[25] Xiao, J., Hu, Y., Xiao, Y., Xu, L., & Wang, S. (2017). A hybrid transfer learning model for crude oil price forecasting. Statistics and Its Interface, 10(1), 119–130. doi:10.4310/sii.2017.v10.n1.a11

[26] Xin, B., & Peng, W. (2020). Prediction for Chaotic Time Series-Based AE-CNN and Transfer learning. Complexity, 2020, 1–9. doi:10.1155/2020/2680480

[27] Lara-Benítez, P., Carranza-García, M., & Riquelme, J. C. (2021). An Experimental Review on Deep learning Architectures for Time Series Forecasting. International Journal of Neural Systems, 31(03), 2130001. doi:10.1142/s0129065721300011

[28] Shynkevich, Y., McGinnity, T. M., Coleman, S. A., Belatreche, A., & Li, Y. (2017). Forecasting price movements using technical indicators: Investigating the impact of varying input window length. Neurocomputing, 264, 71–88. doi:10.1016/j.neucom.2016.11.095

[29] Herbold, S. (2020). Autorank: A Python package for automated ranking of classifiers. Journal of Open Source Software, 5(48), 2173. doi:10.21105/joss.02173

[30] Spiliotis, E., Kouloumos, A., Assimakopoulos, V., & Makridakis, S. (2020). Are forecasting competitions data representative of the reality? International Journal of Forecasting, 36(1), 37–53. doi:10.1016/j.ijforecast.2018.12.007

[31] Wang, X., Smith, K., & Hyndman, R. (2006). Characteristic-Based Clustering for Time Series Data. Data Mining and Knowledge Discovery, 13(3), 335–364. doi:10.1007/s10618-005-0039-x

[32] Box, G. E. P., & Pierce, D. A. (1970). Distribution of Residual Autocorrelations in Autoregressive-Integrated Moving Average Time Series Models. Journal of the American Statistical Association, 65(332), 1509–1526. doi:10.1080/01621459.1970.10481180

[33] Chen, C., & Liu, L.-M. (1993). Forecasting time series with outliers. Journal of Forecasting, 12(1), 13–35. doi:10.1002/for.3980120103

[34] Said, S. E., & Dickey, D. A. (1984). Testing for unit roots in autoregressive-moving average models of unknown order. Biometrika, 71(3), 599–607. doi:10.1093/biomet/71.3.599

[35] Vaiciukynas, E., Danenas, P., Kontrimas, V., & Butleris, R. (2020). Meta-Learning for Time Series Forecasting Ensemble. *arXiv preprint arXiv:2011.10545*.

[36] Oreshkin, B. N., Carpov, D., Chapados, N., & Bengio, Y. (2020). Meta-learning framework with applications to zero-shot time-series forecasting. *arXiv preprint arXiv:2002.02887*.

[37] Li, Y., Zhang, S., Hu, R., & Lu, N. (2021). A meta-learning based distribution system load forecasting model selection framework. Applied Energy, 294, 116991. doi:10.1016/j.apenergy.2021.11699

[38] Lemke, C., & Gabrys, B. (2010). Meta-learning for time series forecasting and forecast combination. Neurocomputing, 73(10-12), 2006–2016. doi:10.1016/j.neucom.2009.09.020.

[39] Makridakis and Hibon (2000) The M3-competition: results, conclusions and implications. International Journal of Forecasting, 16, 451-476.

[40] M4 Team (2018). M4 competitor's guide: prizes and rules. See https:// www.m4.unic.ac.cy/wp-content/uploads/2018/03/M4-CompetitorsGuide.pdf.

[41] Wen, Q., Zhou, T., Zhang, C., Chen, W., Ma, Z., Yan, J., & Sun, L. (2022). Transformers in time series: A survey. arXiv preprint arXiv:2202.07125.